\documentclass[10pt,twocolumn,letterpaper]{article}

\usepackage[pagenumbers]{iccv} 

\usepackage{array}
\newcolumntype{C}[1]{>{\centering\arraybackslash}p{#1}}
\usepackage{bm}
\usepackage{float}
\usepackage{amsmath}
\usepackage{pifont}
\newcommand{\cmark}{\ding{51}}
\newcommand{\xmark}{\ding{55}}
\usepackage{graphicx}
\newcommand{\norm}[1]{\left\lVert#1\right\rVert}
\usepackage[accsupp]{axessibility}  
\usepackage{relsize}

\definecolor{iccvblue}{rgb}{0.21,0.49,0.74}
\usepackage[pagebackref,breaklinks,colorlinks,allcolors=iccvblue]{hyperref}

\def\paperID{9966} 
\def\confName{ICCV}
\def\confYear{2025}

\title{STaR: Seamless Spatial-Temporal Aware Motion Retargeting \\ with Penetration and Consistency Constraints}

\author{
Xiaohang Yang \;\;\;\;\; Qing Wang \;\;\;\;\; Jiahao Yang \;\;\;\;\; Gregory Slabaugh \;\;\;\;\; Shanxin Yuan\\
Queen Mary University of London, United Kingdom\\
{\tt\small \{xiaohang.yang, qing.wang, jiahao.yang, g.slabaugh, shanxin.yuan\}@qmul.ac.uk}
}

\begin{document}
\maketitle

\renewcommand\thefootnote{}%
\let\origFootnotemark\footnotemark  
\let\origFootnotetext\footnotetext  
\footnotetext{\smaller[1]{For the purpose of open access, the author has applied a Creative Commons Attribution (CC-BY) license to any Author Accepted Manuscript version arising.}}

\begin{abstract}

Motion retargeting seeks to faithfully replicate the spatio-temporal motion characteristics of a source character onto a target character with a different body shape. Apart from motion semantics preservation, ensuring geometric plausibility and maintaining temporal consistency are also crucial for effective motion retargeting. However, many existing methods prioritize either geometric plausibility or temporal consistency. Neglecting geometric plausibility results in interpenetration, while neglecting temporal consistency leads to motion jitter.
In this paper, we propose a novel sequence-to-sequence model for seamless \textbf{S}patial-\textbf{T}emporal \textbf{a}ware motion \textbf{R}etargeting (\textbf{STaR}), 
with penetration and consistency constraints.
STaR consists of two modules: (1) a spatial module that incorporates dense shape representation and a novel limb penetration constraint to ensure geometric plausibility while preserving motion semantics, and (2) a temporal module that utilizes a temporal transformer and a novel temporal consistency constraint to predict the entire motion sequence at once while enforcing multi-level trajectory smoothness.
The seamless combination of the two modules helps us achieve a good balance between the semantic, geometric, and temporal targets.
Extensive experiments on the Mixamo and ScanRet datasets demonstrate that our method produces plausible and coherent motions while significantly reducing interpenetration rates compared with other approaches.
Code page: \url{https://github.com/XiaohangYang829/STaR}.

\end{abstract}    
\section{Introduction}
\label{sec:intro}

Motion retargeting is a fundamental task in computer vision and graphics. It involves transferring motion from one character to another—preserving the source character’s dynamics and style while adapting to the target character’s unique structure and geometric constraints. It is widely employed in film production, gaming, and robotic animation. With the rise of digital avatars in virtual environments, rapid advancements in motion generation and digital avatars have further intensified research interest in motion retargeting, which bridges motion capture and character animation. As the demand for authentic visual effects grows, expectations for generated motion have become more stringent. Issues like geometric penetration and unnatural temporal jitter can significantly undermine realism.

\begin{figure}[t]
  \centering
   \vspace{-3mm}
   \includegraphics[width=1.0\linewidth]{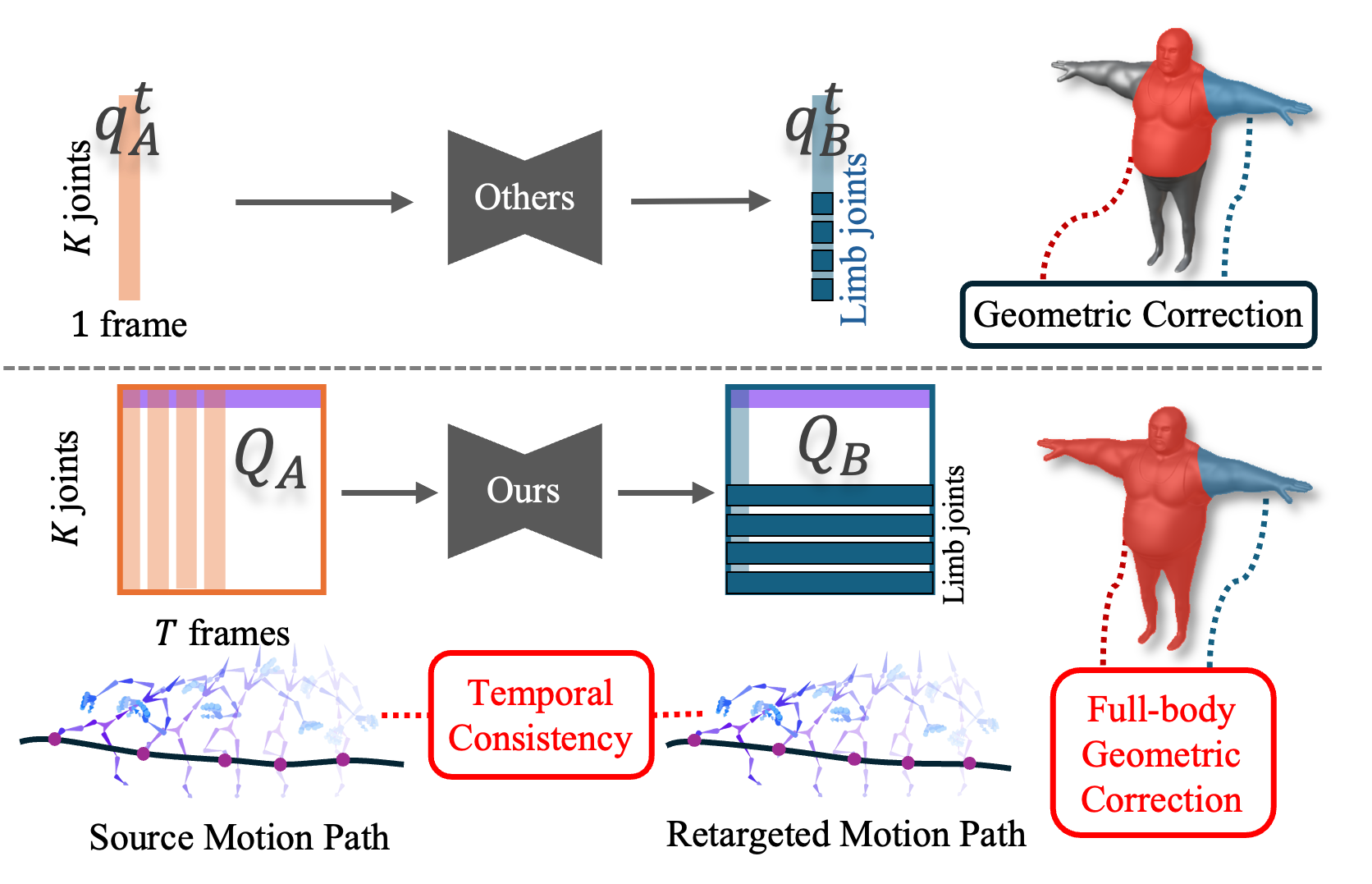}
   \vspace{-5mm}
   \caption{STaR performs sequence-to-sequence motion retargeting, guided by a novel limb penetration constraint and a temporal consistency constraint. Unlike existing methods that focus on limb-body geometric correction, our limb penetration constraint provides full-body geometric correction. The temporal constraint module enhances the smoothness of the entire motion trajectory. STaR effectively reduces the penetration rate while preserving motion semantics and temporal coherence.} 
   \label{fig:intro}
\end{figure}

Motion inherently carries information in both its spatial and temporal dimensions. The spatial relationships among skeletal joints in each motion frame convey critical information about pose semantics, while their temporal evolution defines the motion trajectory. However, many existing methods primarily focus on the spatial domain, treating motion retargeting as a pose transfer problem. Previous approaches \cite{nkn, pmnet, san} generally optimize sparse skeletal rotations and use Forward Kinematics (FK) to derive joint positions, ensuring constraints on both. More recent models \cite{contact, r2et, moma, smt, meshret} extend this by incorporating skinned motion representations to better preserve semantics and mitigate penetration. Although some methods \cite{nkn, pmnet, san, r2et, smt} introduce simple smoothness losses between consecutive frames to improve temporal consistency, temporal glitches persist. Furthermore, neglecting temporal consistency when handling penetration \cite{r2et} can further disrupt motion trajectories.
As illustrated in \cref{fig:intro}, existing methods primarily perform motion retargeting frame-by-frame, often overlooking either geometric correction or temporal consistency.
We propose to ensure geometric plausibility and temporal consistency across the whole sequence through a sequence-to-sequence design.

To minimize penetration and mitigate temporal artifacts in retargeted motion, we propose a novel motion retargeting network that integrates both spatial and temporal supervision, which consists of two key modules: (1) a spatial module that processes skeletal rotations across all joints at each time step to capture motion semantics and geometric cues; (2) a temporal module that models motion dynamics to ensure smooth and coherent motion transitions over time.


In the spatial module, we first design dense shape representations to align the input with dense spatial supervisions, such as penetration loss \cite{r2et, smt}. Instead of solely relying on sparse joint rotations, we densify the input by sampling vertices from both the source and target characters.
Although existing SDF-based penetration loss can effectively improve geometric plausibility, it presents challenges during training. Since each pose corresponds to a unique SDF, obtaining accurate dynamic SDFs is computationally demanding.
To address this, we introduce the limb penetration constraint module, which considers both limb-body penetration and limb-limb penetration. Inspired by the distance function used in point cloud registration \cite{deng2021robust}, we calculate the SDF value for each limb vertex based on the Chamfer Distance. We implemented the limb penetration loss with CUDA \cite{cuda}, allowing us to penalize penetrated vertices efficiently.


To enhance temporal smoothness, we integrate multi-level temporal supervision derived from the source motion. The short-term supervision constrains frame-wise motion directions and magnitudes, while the long-term supervision maintains the overall structure of the motion trajectory. This multi-level temporal guidance helps preserve motion smoothness and coherence.
To evaluate STaR's performance, we conduct extensive experiments across various scenarios, including unseen motion and characters, and the motion from real humans. Both qualitative and quantitative analyses demonstrate that our method outperforms existing approaches, achieving more natural, coherent, and physically plausible results.




Our contributions are summarized as follows:
\begin{itemize}
\item We introduce a novel spatio-temporal network for motion retargeting that concurrently processes pose spatial information and temporal dynamics.
\item We apply dense shape representations and propose limb penetration loss to avoid penetration while preserving motion semantics.
\item We develop a temporal consistency loss to ensure motion smoothness, which effectively reduces motion jitter.
\item We validate our method through comprehensive experiments on Mixamo \cite{mixamo} and ScanRet \cite{meshret} datasets.
\end{itemize}

\section{Related Work}
\label{sec:relatedworks}

\textbf{Optimization-based Motion Retargeting.}
Early studies \cite{retargettingretargetting, lee1999hierarchical, choi2000online, tak2005physically, feng2012automating} explored optimization-based motion retargeting using manually designed kinematic constraints and assumptions. Gleicher \cite{retargettingretargetting} introduces a space-time constraint solver to preserve key motion features while adapting movement to characters with different proportions. Lee and Shin \cite{lee1999hierarchical} combine hierarchical curve fitting with an inverse kinematics solver to smoothly refine motion using multilevel B splines. Choi and Ko \cite{choi2000online} develop a real-time motion adaptation technique that dynamically adjusts joint angles for interactive applications. Unlike traditional spacetime optimization, Tak and Ko \cite{tak2005physically} propose a per-frame Kalman filter framework for stable, physically plausible motion retargeting. Feng \etal \cite{feng2012automating} automate skeleton mapping and motion retargeting, enabling fast character integration. While effective in controlled scenarios, they are constrained by limited training data and simplified motion models.

\noindent \textbf{Neural Motion Retargeting.} With the advent of large motion datasets and deep learning, neural motion retargeting methods have gained popularity in recent years. Villegas \etal \cite{nkn} introduce a recurrent neural network with a Forward Kinematics (FK) layer and a cycle consistency-based adversarial training objective to perform unsupervised motion retargeting. Lim \etal \cite{pmnet} separate pose retargeting and movement adaptation into two distinct processes, and achieve natural motion transfer. Aberman \etal \cite{san} enable unpaired motion retargeting across skeletons with different structures by embedding motion into a common latent space using differentiable skeleton-aware convolution, pooling, and unpooling operators. Li \etal \cite{li2022iterative} propose a deep learning-based iterative method for motion retargeting that avoids adversarial training and instead uses an autoencoder-based latent space optimization to improve generalization to unseen characters and motions.
All these methods \cite{nkn, pmnet, san, li2022iterative} above are based on the skeletal motion, while \cite{contact, r2et, smt, meshret} consider body geometry.
Villegas \etal \cite{contact} employ a geometry-conditioned recurrent network and encoder-space optimization to preserve self-contacts and prevent interpenetration. Zhang \etal \cite{r2et} introduce a residual-based neural retargeting framework that progressively adjusts source motions to fit target skeletons and shapes, and use distance fields to avoid interpenetration and ensure self-contact. Zhang \etal \cite{smt} design a motion retargeting framework based on Vision Language Model that extracts and preserves high-level motion semantics using differentiable rendering. Martinelli \etal \cite{moma} use a transformer-based masked autoencoder for skeletal adaptation and a face-based optimization approach to prevent mesh interpenetration. Zhang \etal \cite{meshret} directly models dense mesh interactions using semantically consistent sensors (SCS) and a Dense Mesh Interaction (DMI) field, enabling accurate motion transfer.
Based on these works, our method further enhances the physical plausibility and simultaneously improves the motion smoothness to a level nearly indistinguishable from the source.

\section{Methods}

\begin{figure}[t]
  \centering
   \includegraphics[width=1.0\linewidth]{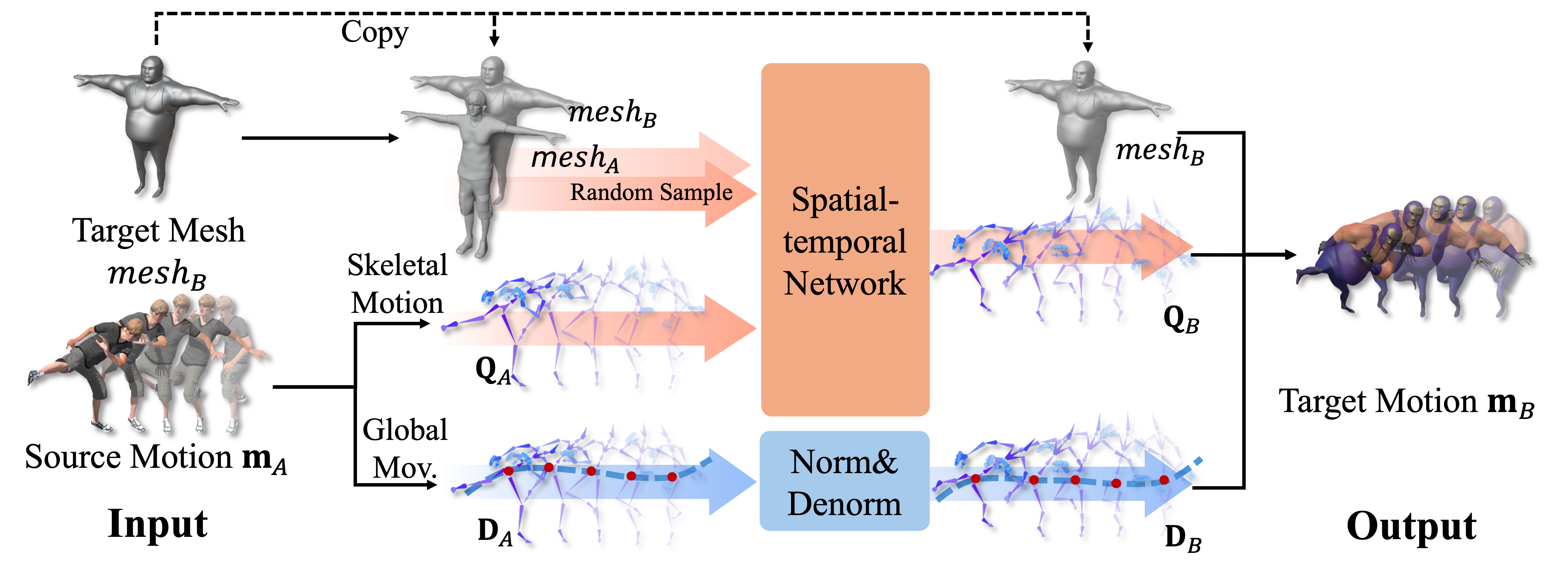}
   \caption{Overview of STaR, which splits the input motion into two modalities: skeletal motion sequence, and global movements. STaR also takes the meshes of the source and target character as input when retargeting the motion.
   }
   \label{fig:abstract}
\end{figure}

The overall framework of STaR is introduced in ~\cref{network}. Since motion retargeting lacks paired ground truth, we leverage additional supervision signals from the source motion sequences: a limb penetration constraint module (\cref{spatial}) to avoid interpenetration and a temporal consistency module (\cref{temporal}) to maintain motion coherence. The overall loss function, integrating these constraints with semantic supervision, is described in~\cref{loss}.


\subsection{Spatio-Temporal Aware Motion Retargeting} \label{network}

As shown in \cref{fig:abstract}, given a source motion $\mathbf{m}_A$ from the source character $A$ and the T-posed meshes of the source character ${mesh}_A$, and target character ${mesh}_B$, the goal is to generate the target motion $\mathbf{m}_B$ while preserving motion semantics, geometric plausibility, and temporal consistency. To achieve this, we decompose the input $\mathbf{m}_A$ into skeletal motion, represented by rotation quaternions $\mathbf{Q} \in \mathbb{R}^{T \times K \times 4}$, and global movement $\mathbf{D} \in \mathbb{R}^{T \times 4}$, capturing both local skeletal rotations and the root joint's trajectory. Here, $T$ denotes the length of the motion sequence, and $K$ represents the number of skeletal joints.
Because characters vary in height, we compensate for this by normalizing global motion, then restoring its original scale afterward.


At each iteration, we randomly sample a subset of points from ${mesh}_A$ and ${mesh}_B$, resulting in two point sets $\mathbf{P}_A$ and $\mathbf{P}_B$. 
These point sets ($\mathbf{P}_A$ and $\mathbf{P}_B$), along with the skeletal motion ($\mathbf{Q}_A$), are fed into the spatio-temporal model, which predicts a residual motion adjustment ($\Delta \mathbf{Q}$) that aligns the retargeted motion with the target character’s structure. The complete framework can be formulated as:  

\vspace{-3mm}
\begin{equation} \label{eq1}
\Delta \mathbf{Q} = \mathcal{F} ( \mathbf{Q}_A, \mathbf{P}_A, \mathbf{P}_B ), \  \mathbf{Q}_B = \Delta \mathbf{Q} \otimes \mathbf{Q}_A,
\end{equation}
where $\mathcal{F}$ represents the spatio-temporal model.

\begin{figure*}[ht]
  \centering
   \includegraphics[width=1.0\linewidth]{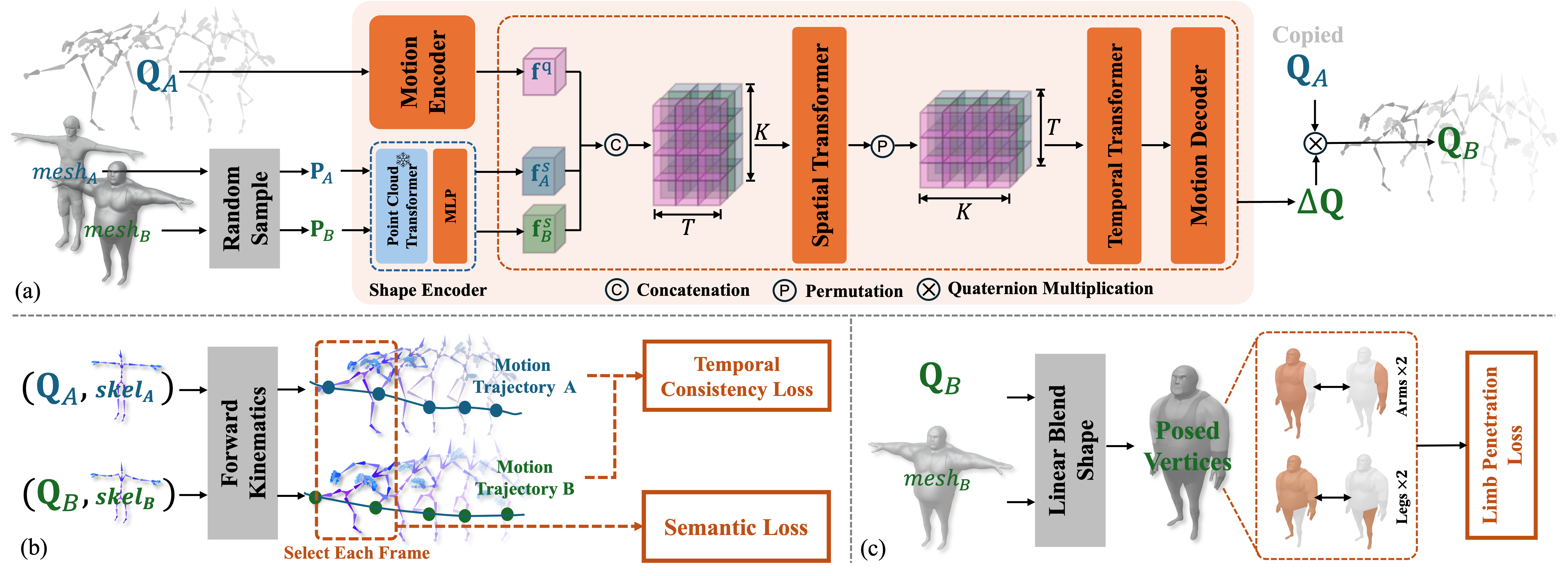}
   \caption{Details of STaR. (a) The detailed architecture of the spatio-temporal model. (b) The constraints on skeletal motion include temporal consistency loss and semantic loss. (c) The limb penetration loss is applied to the skinned motion.}
   \label{fig:detail}
\end{figure*}

\noindent \textbf{Motion Encoder and Shape Encoder.} As shown in \cref{fig:detail} (a), like most retargeting methods \cite{pmnet, r2et, smt}, we utilize a motion encoder to process the skeletal motion $\mathbf{Q}_A$ to motion feature $\mathbf{f}^q \in \mathbb{R}^{T \times K \times C}$, where $C$ is the latent length. However, this sparse input cannot provide dense information for motion retargeting when there are dense supervision signals applied.


To address this, we incorporate a frozen off-the-shelf point cloud network \cite{pct} as a geometric feature extractor. Since most point cloud learning methods require a fixed number of input points, we randomly sample point sets from the T-posed meshes of both source and target characters at each iteration. Because the sampled points vary in each iteration, the network is exposed to different geometric subsets over time, forcing it to develop a more robust shape awareness rather than overfitting to specific sampled regions. This strategy enhances the network’s ability to generalize across different body shapes while ensuring a consistent understanding of detailed geometry.

The output of the dense shape encoder is further processed by a small multi-layer perceptron (MLP), which takes the logits from the point cloud transformer as input and generates a shape feature $\mathbf{f}^s \in \mathbb{R}^{T \times K \times C}$. This feature is designed to align with the representation produced by the motion encoder, ensuring compatibility between shape and motion features. It is particularly beneficial for subsequent spatio-temporal operations, as it enables smooth information exchange across both spatial and temporal dimensions, improving the model’s ability to learn coherent motion transformations.  

\noindent \textbf{Spatio-Temporal Network.} The motion feature \(\mathbf{f}^q\) from the motion encoder and the dense shape representations \(\mathbf{f}^s_A\) and \(\mathbf{f}^s_B\) are first concatenated along the latent dimension, forming a unified representation of size \(\mathbb{R}^{T \times K \times (3*C)}\). This combined feature is then processed by the spatial transformer, which captures spatial relationships between skeletal joints at each time step, ensuring structural consistency and geometric plausibility.

After spatial processing, we apply a permutation operation, swapping the temporal and spatial dimensions to reshape the input from \(\mathbb{R}^{T \times K \times (3*C)}\) to \(\mathbb{R}^{K \times T \times (3*C)}\). This step enables the temporal transformer to model motion evolution for each joint individually, focusing on temporal consistency. For each joint, the temporal transformer extracts temporal attention across its motion trajectory, capturing both local smoothness and long-term dependencies. This ensures coherent motion transitions while maintaining temporal consistency throughout the sequence. Finally, the motion decoder predicts a residual motion correction, which is applied to the source motion \(\mathbf{Q}_A\) to obtain the retargeted skeletal motion \(\mathbf{Q}_B\).

\subsection{Limb Penetration Constraint Module} \label{spatial}

\begin{figure}[t]
  \centering
   \includegraphics[width=1.0\linewidth]{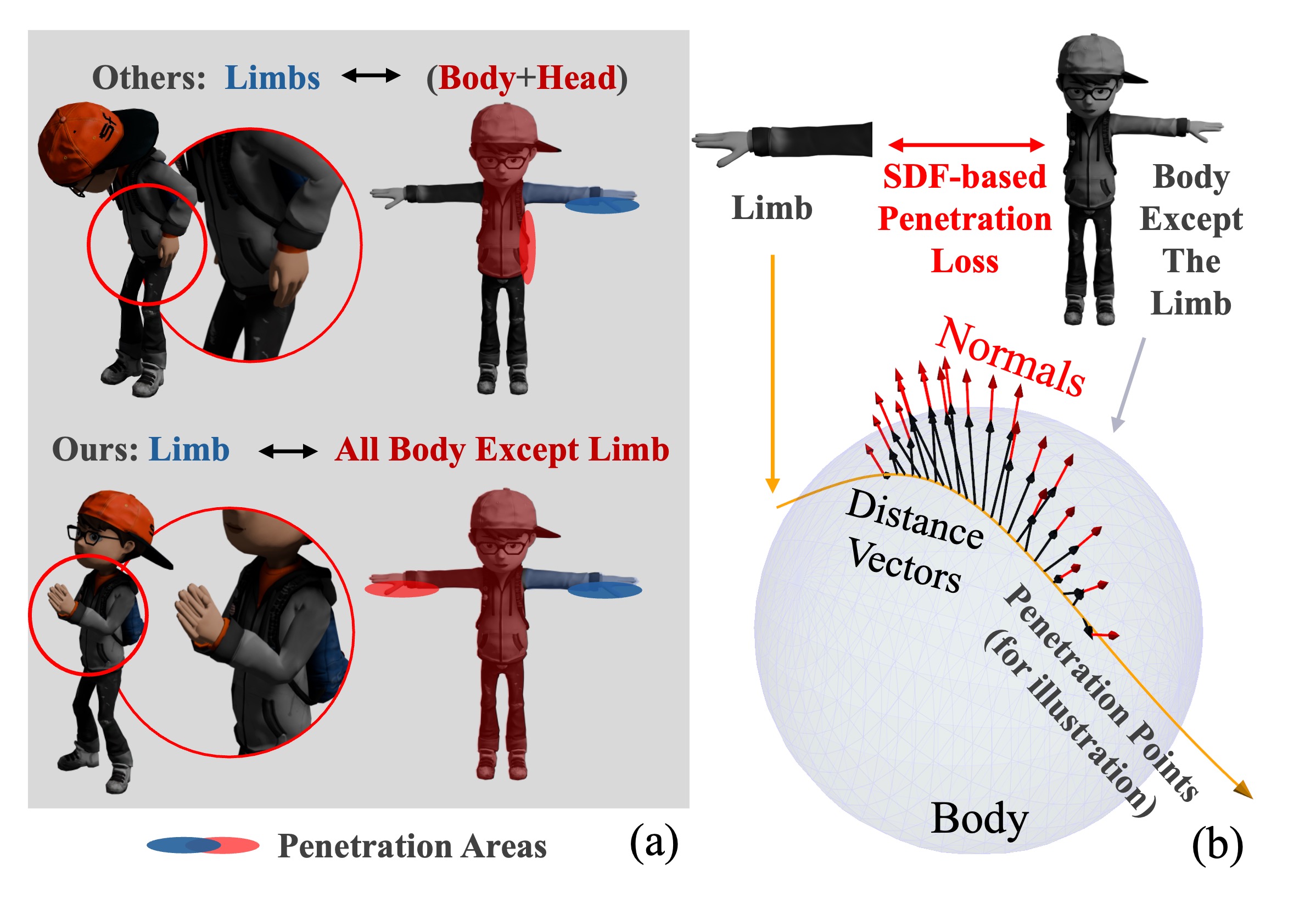}
   \vspace{-6mm}
   \caption{Illustration of different types of interpenetration. Existing methods \cite{r2et} focus on the interpenetration between limbs and body or head. While for motions like clapping, two hands tend to penetrate each other. We design a novel normal-based SDF loss to include all kinds of interpenetration.}
   \label{fig:penetration}
\end{figure}

To prevent penetration, we propose \textit{limb penetration loss}. As illustrated in \cref{fig:penetration} (a), this loss is designed based on the observation that penetration occurs not only between the limbs and the main body \cite{r2et}, but also \textit{between} the limbs. To mitigate penetration, we penalize each limb's penetration with all other body parts, as illustrated in \cref{fig:detail} (c).

The calculation of the \textit{limb penetration loss} is inspired by Chamfer Distance and point cloud registration \cite{deng2021robust}, which are based on measuring the distance between two point sets. As clarified in \cref{fig:penetration}, the loss calculation follows these steps (1) deforming the T-posed vertices and normal vectors of the target character according to $Q_B$ using Linear Blend Skinning (LBS) to obtain posed vertices for the limb and the rest of the body, serving as query and reference vertices, respectively; (2) computing the Chamfer Distance between the query and reference vertices; (3) identifying the nearest reference vertex for each query vertex; and (4) multiplying the vector from the query vertex to its nearest reference vertex by the deformed normal vector of the reference vertex, yielding the penetration loss for that query vertex.
Specifically, for each query vertex $\mathbf{e}$ located on the limbs, the nearest reference vertex $\mathbf{e}^r$ with normal vector $\mathbf{n}^r$ is identified using a Nearest Neighbor Search (NNS) algorithm. The SDF value for the query vertex $\phi(\mathbf{e})$ is calculated by the dot product $\mathbf{v}^{r} \cdot \mathbf{n}^r$, where $\mathbf{v}^{r} = \mathbf{e} - \mathbf{e}^r$. The overall penetration penalty $\mathcal{L}_{lp}$ is defined as follows:

\vspace{-2mm}
\begin{equation} \label{eq7}
\mathcal{L}_{lp} = \frac{1}{\sum_{l} N_l} \sum_{l} \sum_{\mathbf{e} \in \mathbf{E}_l} ReLU\left(\phi(\mathbf{e})\right),
\end{equation}
where $l$ is one of the four limbs (right arm, left arm, right leg, left leg), $N_l$ is the vertex number for each limb, and the vertex set $\mathbf{E}_l \!= \{ \mathbf{e}_i \}_{i=1}^{N_l}$. The reference vertex set is $\mathbf{E}_l^r \!= \{ \mathbf{e}_j^r \}_{j=1}^{N_l^r}$. Sampling the same points for all characters within a batch allows us to process the limb penetration loss for the entire batch simultaneously. In our model, we assume that the SDF value is positive when a vertex is inside the body. We apply the $ReLU$ function to retain all positive SDF values, ensuring that only penetrating vertices contribute to the penetration loss. Additionally, since the LBS algorithm introduces partial interpenetration, we exclude certain body parts, such as the upper arm, from the loss computation.

Please refer to \cref{penetration} in the supplemental materials for the detailed image, which displays the conditions of both penetrating vertices and non-penetrating ones.

\subsection{Temporal Consistency Constraint Module} \label{temporal}

\begin{figure}[t]
  \centering
   \includegraphics[width=1.0\linewidth]{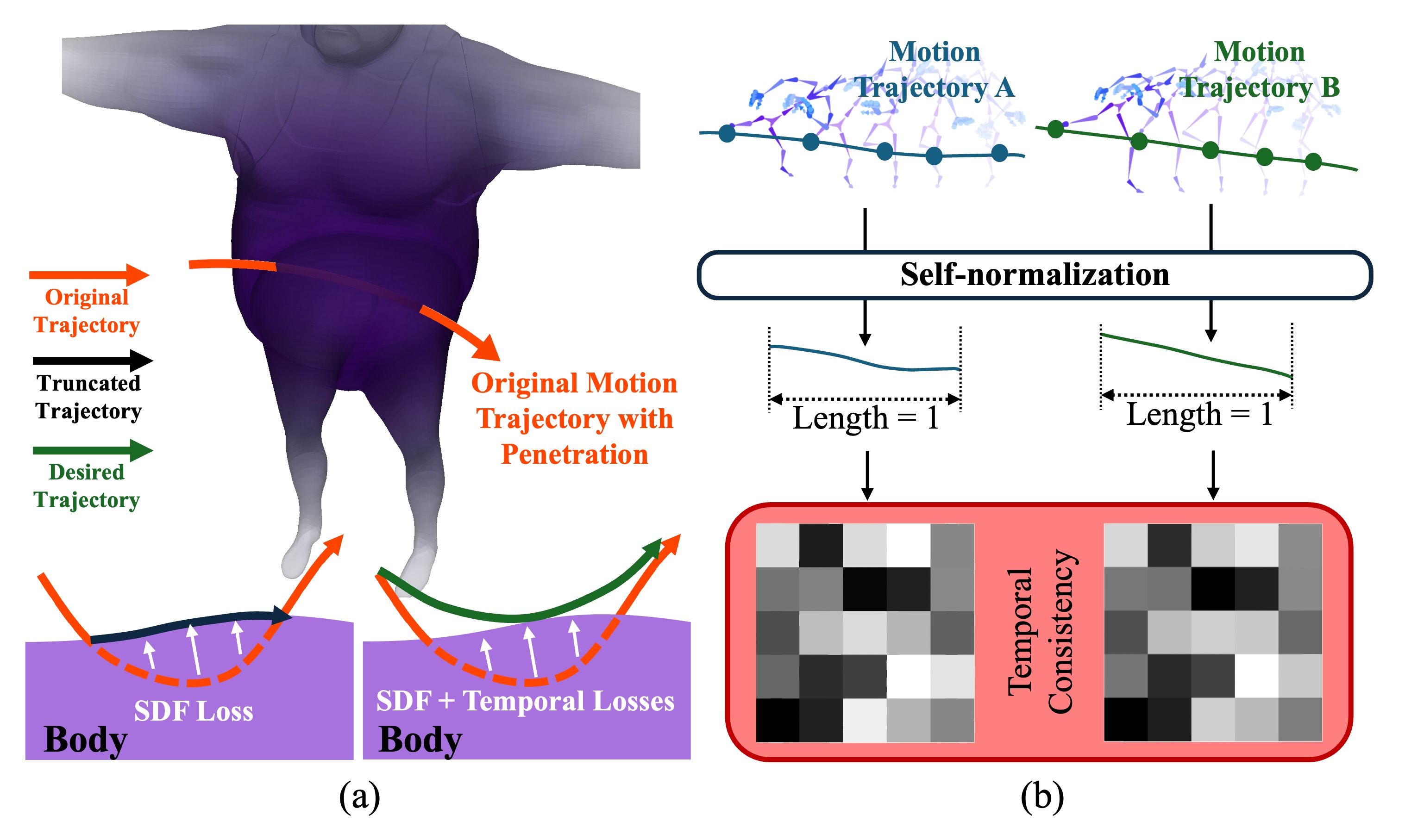}
   \vspace{-6mm}
   \caption{(a): Geometric correction for motion path with penetration will lead to motion jitter. Motion smoothness should be retained while avoiding penetration. The green curve is the desired motion path, balancing penetration avoidance and motion coherence. (b): Illustration of the temporal consistency. The motion paths from different characters are self-normalized to eliminate the effect of different skeleton structures.}
   \label{fig:temporal}
\end{figure}

As illustrated in \cref{fig:temporal} (a), the motion path may intersect with the body, causing penetration. When applying SDF loss, the spatial module often truncates the motion path to prevent penetration, disrupting its smoothness and contributing to temporal jitter. Ideally, instead of abrupt truncation, the motion path should be warped adaptively based on penetration conditions along its trajectory—even in regions where no penetration occurs. To achieve this, we introduce a \textit{temporal transformer} \cite{transformer} to enable information exchange across different time frames. Additionally, we propose a multi-level \textit{temporal consistency loss} that constrains both short-term motion vectors and long-term motion trajectories, ensuring a smoother and more coherent motion path.

Simple smoothness losses used by other works focus on consecutive frames, whose receptive field is quite limited and fails to perceive the smoothness of the full motion path. To narrow the consistency gap between the generated motion and the source motion, we introduce a multi-level \textit{temporal consistency loss}. We obtain the motion trajectory of each joint by manipulating the skeleton with Forward Kinematics (FK) according to the predicted skeletal rotations $\mathbf{Q}_B$, as demonstrated in \cref{fig:detail} (b). Unlike \cite{smt}, we do not normalize the motion by the character height or skeleton length. Instead, we normalize the trajectory by itself, scaling it to a unit cube to neglect the influence of different skeleton structures, as displayed in \cref{fig:temporal} (b). We maintain the motion dynamics by constraining the pair-wise motion vectors. For the joint $k$, the normalized trajectory is $\{\mathbf{c}^k_t\}_{t=1}^T$. For each point pair $(\mathbf{c}^k_{i}, \mathbf{c}^k_{j}), i, j \in \{ 1, ... T\}$, we calculate the motion vector $\mathbf{m}_{i,j}^k$ from the start point $\mathbf{c}^k_{i}$ to the endpoint $\mathbf{c}^k_{j}$. Stacking the motion vectors together, we create a matrix $\mathbf{M}^k \!\in \mathbb{R}^{T\times T\times3}$.
As the \cref{fig:temporal} (b) shows, we use the difference of the motion matrix $\mathbf{M}^k$ in the \textit{temporal consistency loss}. The motion matrix is calculated for each joint of the source motion, and it acts as the supervision of the motion dynamics of its corresponding joint in the retargeted motion. The \textit{temporal consistency loss} is defined as:

\vspace{-2mm}
\begin{equation} \label{eq8}
\mathcal{L}_{tc} = \frac{1}{K}\sum_{k=1}^K\norm{\mathbf{M}_A^k - \mathbf{M}_B^k}^2_2.
\end{equation}

\subsection{Loss Functions} \label{loss}

As illustrated in \cref{fig:detail}, unlike previous multi-stage models \cite{r2et, smt}, we train STaR in an end-to-end manner. And as there are no paired datasets for the motion retargeting task, we exploit as many supervision signals as possible from the source motion to achieve unsupervised learning. Apart from the limb penetration loss and temporal consistency loss introduced in \cref{spatial} and \cref{temporal}, we utilize several semantic losses to preserve the motion semantics.

\noindent \textbf{Reconstruction Loss.} We adopt a self-reconstruction strategy following \cite{pmnet, r2et}. The objective of self-reconstruction is to constrain the retargeting of the source motion $\mathbf{Q}_A$ onto the source character $A$. This strategy assists the network in searching for the static points of the neural function and ensures the stabilization of the model during training. The reconstruction loss $\mathcal{L}_{rec}$ includes skeletal rotation reconstruction and joint position reconstruction, which is defined as:

\vspace{-1mm}
\begin{equation} \label{eq9}
\begin{aligned}
\mathcal{L}_{rec} = \ & \lVert\mathbf{Q}_A - \hat{\mathbf{Q}}_A\rVert^2_2 \  + \\
&\lVert  f_{K}(\mathbf{Q}_A,\gamma_A) - f_{K}(\hat{\mathbf{Q}}_A,\gamma_A)  \rVert^2_2,
\end{aligned}
\end{equation}
in which $\hat{\mathbf{Q}}_A$ is the reconstructed motion. $f_K$ is the Forward Kinematics (FK) function, which projects the T-posed skeleton $\gamma\!\in\!\mathbb{R}^{K\times3}$ to the specific pose.






\noindent \textbf{Constraint Loss.} In contrast to the reconstruction loss, while retargeting motion to the target character $B$, we use the constraint loss to limit modification while reducing penetration. To constrain the modification as small as possible, we encourage the retargeted motion $\mathbf{Q}_B$ to be close to the source motion $\mathbf{Q}_A$:

\vspace{-1mm}
\begin{equation} \label{eq13}
\begin{aligned}
\mathcal{L}_{con} = \ & \lVert \mathbf{Q}_B - \mathbf{Q}_A\rVert^2_2 \ + \\
& \lVert  f_{K}(\mathbf{Q}_B,\gamma_B) - f_{K}(\mathbf{Q}_A,\gamma_B)  \rVert^2_2,
\end{aligned}
\end{equation}
where $\gamma_B$ is the T-posed skeleton of the target character, and $f_{K}(\mathbf{Q}_A,\gamma_B)$ is the joint position after copying motion $\mathbf{Q}_A$ to the target character $B$.

\noindent \textbf{Joint Orientation Loss.} Under the joint influence of multiple losses, the network sometimes predicts motion that appears consistent with the source motion but results in an inward-outward inversion of the arms. In such cases, while the skeletal motion and penetration rate remain reasonable, the semantic meaning of the action is altered. To address this, we bind vectors $\mathbf{J}$ to all skeletal joints, which transforms along with the Forward Kinematics process. By monitoring the direction of this vector, we ensure that each skeletal joint maintains a reasonable orientation. The joint orientation loss is formulated as:

\vspace{-2mm}
\begin{equation} \label{eq14}
\mathcal{L}_{j} = \lVert f_{K}(\mathbf{Q}_A, \mathbf{J}) - f_{K}(\mathbf{Q}_B, \mathbf{J})\rVert^2_2,
\end{equation}
where $\mathbf{J}$ is identical for every character.

The overall loss function is:

\vspace{-5mm}
\begin{equation} \label{eq15}
\mathcal{L}_{all} = \lambda_{rec} \mathcal{L}_{rec} + \lambda_{con} \mathcal{L}_{con} + \lambda_{lp} \mathcal{L}_{lp} + \lambda_{tc} \mathcal{L}_{tc} + \lambda_{j} \mathcal{L}_{j}.
\end{equation}

\noindent \textbf{Global motion.} According to our observation, the dataset \cite{mixamo} lacks sufficient diversity to effectively train a stable global translation decoder, particularly for predicting translations of unseen characters. The neural function will overfit the training set. Thus, we follow the normalization method from \cite{r2et}.

\section{Experiments}

\noindent \textbf{Datasets.} We train and evaluate STaR, along with comparative methods, on the Mixamo dataset \cite{mixamo}, a collection of diverse motion clips performed by artificial characters with various skeleton configurations and body shapes. We collect 2,400 motion instances from 11 characters, splitting them into training and testing sets. The training set comprises 1,905 motion sequences, with source and target characters sampled from a pool of 7 characters per sequence. Each source motion is randomly sampled for 60 frames to facilitate motion retargeting training.
For testing, we follow the evaluation setup of previous works \cite{nkn, pmnet, san, r2et}. The test set includes both Seen and Unseen Characters (SC, UC), as well as Seen and Unseen Motion clips (SM, UM), forming four subsets: SC+SM, UC+SM, SC+UM, and UC+UM. Only SC+SM is encountered during training. We sample 100 pairs per subset, segment each motion into 120-frame sequences, and report final results as the average across all subsets.
To assess performance on real-human motion retargeting, we also test on the ScanRet dataset \cite{meshret}, which contains 8,298 motion clips from 100 human actors. During inference, we follow the official split from \cite{meshret}, consisting of 9 motion clips and 10 human actors.

\noindent \textbf{Implementation details.} We select $K = 22$ nodes per character and sample $N = 1024$ vertices for geometric embedding in each forward pass of the network. The feature length is set to $S = 64$. The hyperparameters are configured as follows: $\lambda_{rec} = 0.1$, $\lambda_{con} = 0.1$, $\lambda_{lp} = 5.0$, $\lambda_{tc} = 1.0$, and $\lambda_{j} = 1.0$. The limb penetration loss is partially implemented with CUDA \cite{cuda} for acceleration. We set $N_l$ and $N_l^r$ to $(50, 100, 200, 400)$, and $(500, 1{,}000, 2{,}000, 4{,}000)$ respectively, and the corresponding results can be found in \cref{ablation}. We train the network using the Adam optimizer \cite{adam} and incorporate a frozen point cloud model from PCT \cite{pct}. The entire training process runs for 50 epochs.

\noindent \textbf{Evaluation metrics.} We evaluate the effectiveness of our model across three key aspects: joint accuracy, geometric penetration, and motion smoothness. For joint accuracy, we compute the Mean Squared Error (MSE) and local MSE (MSE${}^{lc}$) to assess how closely the retargeted joint coordinates align with the ground truth (GT). Local MSE isolates local motion accuracy by eliminating the influence of global motion. These errors are normalized by character height. From a geometric perspective, we compare the penetration rate (Pen$\%$), which measures the percentage of limb vertices that penetrate other body parts. Unlike most existing methods \cite{r2et, smt}, which consider only the body and head as penetrable regions, we also account for interpenetration between limbs for a more comprehensive evaluation. Few existing methods have established quantitative metrics for evaluating retargeted motion. We introduce a curvature metric (Curv), which quantifies the smoothness of the discrete motion trajectory of each joint. The final curvature score is computed as the average curvature across all limb joints. Further details on these evaluation metrics are provided in \cref{metrics} of the supplementary materials.

\subsection{Qualitative Results}

\begin{figure*}[ht]
  \centering
   \includegraphics[width=1.0\linewidth]{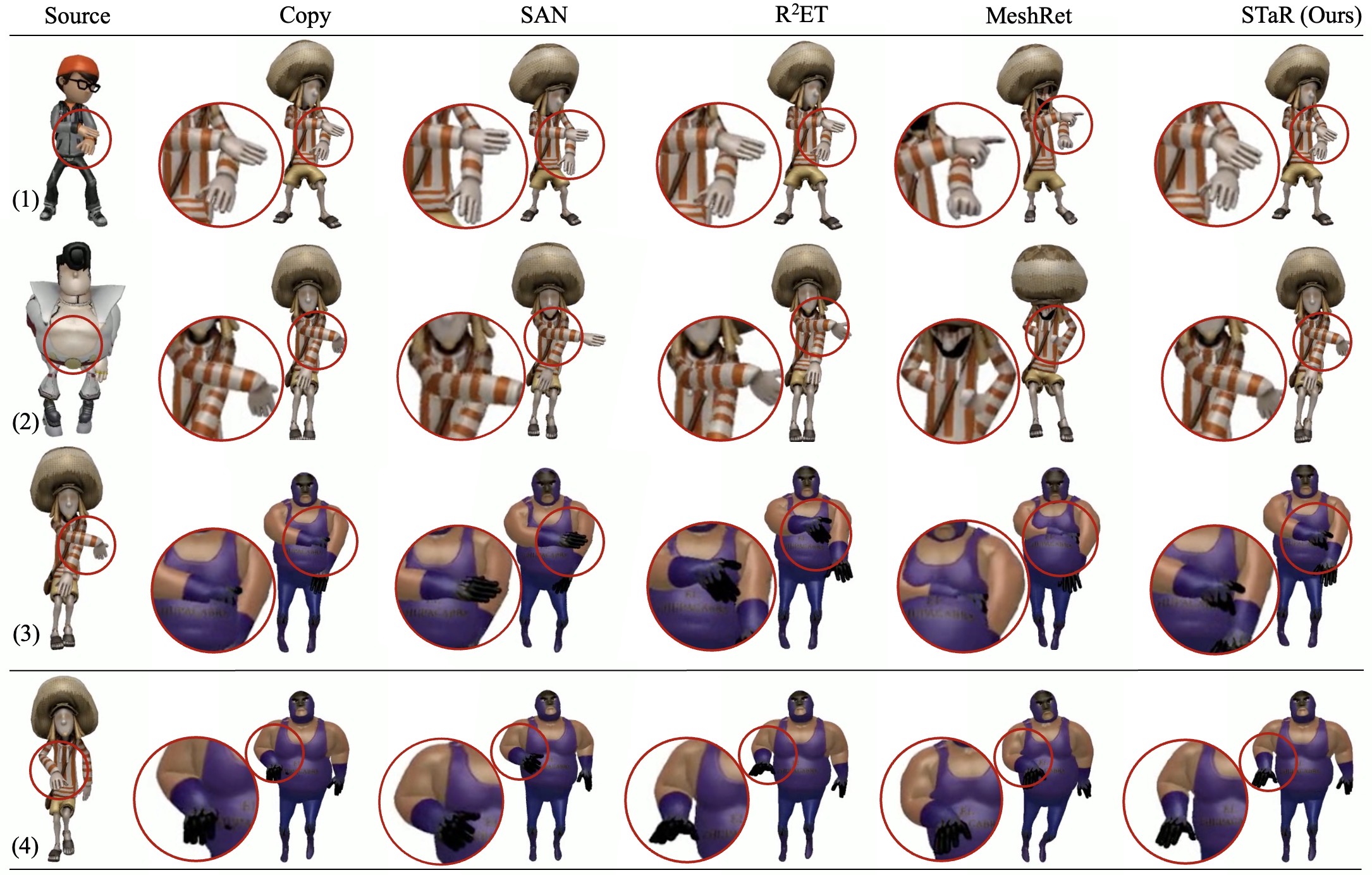}
   \caption{Qualitative comparison with other methods. The top three rows compare motion retargeting across different body shapes: same body shape, fat-to-slim, and slim-to-fat. The bottom row presents a preceding frame relative to the frame of the third row, highlighting the impact of our temporal consistency module on the spatial results. Please ignore hand details.}
   \label{fig:qualitative}
\end{figure*}

We compare our method with ``Copy'', SAN \cite{san}, R${}^{2}$ET \cite{r2et}, and MeshRet \cite{meshret}, and visualize the skinned results in \cref{fig:qualitative}. The first row shows that when the source and target characters have similar body shapes, all methods preserve motion semantics well, but STaR produces results closest to the source motion.
In the second and third rows, when retargeting motion to characters with significantly different body shapes, penetration issues become more pronounced. Copy and SAN \cite{san} struggle to adapt motion naturally when transferring motion from slim to fat, leading to severe penetration. MeshRet \cite{meshret}, though designed for geometric constraints, fails when the source motion already contains interpenetration. R${}^{2}$ET produces competitive results but suffers from severe motion jitter. Despite body shape constraints, STaR effectively preserves motion semantics while significantly reducing penetration artifacts through full-body geometric correction.
The last row, which is the preceding motion frame relative to the third row, partially reflects temporal consistency, showing that STaR produces a more stable and natural motion trajectory. Unlike other methods that aligned frame-wise arm movements, STaR maintains a larger motion space and refines motion trajectories even when no interpenetration occurs, demonstrating its ability to adjust trajectories across the entire sequence.
We further infer STaR on the real-human dataset, ScanRet \cite{meshret}, and the results are shown in \cref{realhuman} in the supplemental materials.



\subsection{Quantitative Results}




\Cref{tab:quantitative} compares our method with current state-of-the-art approaches. The ``Copy'' baseline achieves the lowest MSE and local MSE, since Mixamo \cite{mixamo} may generate dataset samples by copying and slightly modifying motions—yielding artificially low error metrics but at the cost of semantic accuracy and increased interpenetration, especially for characters with differing body shapes, as shown in \cref{fig:qualitative}. Among all motion retargeting methods, R${}^{2}$ET \cite{r2et} achieves the lowest MSE, demonstrating its ability to preserve motion semantics. Our method, STaR, maintains similarly low MSE while adjusting metric weights to prioritize reduced penetration rate and curvature.

Regarding geometric plausibility, skeletal-only retargeting methods such as PM-Net \cite{pmnet} and SAN \cite{san} exhibit high penetration rates—typically exceeding $11\%$—because they disregard character geometry during motion retargeting. While these methods generally preserve motion smoothness. Methods incorporating geometric constraints, such as R${}^{2}$ET \cite{r2et} and MeshRet \cite{meshret}, effectively reduce penetration rate but introduce excessive motion jitter, as indicated by elevated curvature. R${}^{2}$ET${}^{\dag}$ \cite{r2et} addresses this by predicting balancing weights to smoothen the motion, which significantly lowers curvature at the expense of a slight increase in penetration rate. Our method, STaR (Curv), achieves lower penetration rate ($8.70\%$), outperforming R${}^{2}$ET \cite{r2et} while also producing lower curvature than R${}^{2}$ET${}^{\dag}$ \cite{r2et}, with only a slight increase in MSE. During the process of preventing interpenetration, curvature inevitably increases slightly, and this modest increase does not significantly affect motion smoothness. By reducing the penetration rate instead, we further enhance motion plausibility. We relax the smoothness constraint and achieve the lowest penetration rate of $7.99\%$, which is $16.4\%$ lower than R${}^{2}$ET \cite{r2et}. These results demonstrate that STaR effectively balances motion semantics, geometric plausibility, and temporal consistency, highlighting the benefits of our spatio-temporal model, limb penetration constraint, and temporal consistency constraint.

For human perception evaluation, please refer to \cref{userstudy}.

\begin{table}[h]
\caption{Quantitative comparisons with state-of-the-art methods. MSE${}^{lc}$ is the local MSE. R${}^2$ET${}^{\dag}$ \cite{r2et} refers to the debugged version of R${}^2$ET \cite{r2et}. \textbf{STaR} (Curv) assigns a higher weight to motion smoothness. For MSE, MSE${}^{lc}$, and Pen\%, lower values indicate better performance. For Curv, closer to GT's Curv is better.} 
\centering 
\scalebox{0.75}{
\begin{tabular}{l|cc|c|c} 
\toprule[2pt] 
\textbf{Methods} & $\ \textbf{MSE}\downarrow\ $ & $\textbf{MSE}^{lc}\downarrow$ & $\textbf{Pen}\%\downarrow$ & $\textbf{Curv}$ \rotatebox[origin=c]{180}{$\Lsh$} \\ 
\midrule 
Source & - & - & - & 6.54 \\
GT & - & - & 11.26 & 7.64 \\ 
\midrule
Copy & 0.0308 & 0.0102 & 11.23 & \textbf{7.66} \\
\midrule
\multicolumn{5}{c}{\textbf{Skeletal-only motion retargeting methods}}\\
\midrule
PM-Net \cite{pmnet} & 0.2469 & 0.0287 & 11.46 & 12.06\\ 
SAN \cite{san} & 0.1664 & 0.0256 & 12.43 & 8.16 \\ 
\midrule
\multicolumn{5}{c}{\textbf{Skinned motion retargeting methods}}\\
\midrule
R${}^2$ET \cite{r2et} & 0.0335 & 0.0138 & 9.55& 35.67 \\
R${}^2$ET${}^{\dag}$ \cite{r2et} & 0.0331 & 0.0132 & 9.92 & 7.87 \\
MeshRet \cite{meshret} & 0.0691 & 0.0325 & 9.39 & 17.67 \\
\midrule
\textbf{STaR} (Curv) & 0.0353 & 0.0158 & \underline{8.70} & \underline{7.76} \\
\textbf{STaR} (Ours Final) & 0.0368 & 0.0174 & \textbf{7.99} & 10.66 \\
\bottomrule[2pt] 
\end{tabular}
}
\label{tab:quantitative}
\end{table}

\subsection{Ablation Study} \label{ablation}

\begin{table}[h]
\caption{Ablation Study. Baseline is a model which retargets motion frame-by-frame, training with $\mathcal{L}_{sdf}$ from \cite{r2et}. ST is the spatio-temporal model, and DSR stands for dense shape representation. $\mathcal{L}_{lp}$ is the limb penetration constraint, and $\mathcal{L}_{tc}$ is the temporal consistency constraint. $\mathcal{L}_t$ is a basic motion smoothness loss. \cmark\!\!\cmark stands for a higher weight for $\mathcal{L}_{lp}$. In the second part of the table, we sample different numbers of vertices for the limb penetration constraint.} 
\centering 
\scalebox{0.75}{
\begin{tabular}{C{0.7cm}C{0.7cm}C{0.7cm}C{0.7cm}|cc|c|c} 
\toprule[2pt] 
ST & $\mathcal{L}_{lp}$ & DSR & $\mathcal{L}_{tc}$ & $\ \textbf{MSE}\downarrow\ $ & $\textbf{MSE}^{lc}\downarrow$ & $\textbf{Pen}\%\downarrow$ & $\textbf{Curv}$ \rotatebox[origin=c]{180}{$\Lsh$} \\ 
\midrule 
\xmark & $\mathcal{L}_{sdf}$ & \xmark & \xmark & 0.0479 & 0.0321 & 8.74 & 66.35 \\
\cmark & $\mathcal{L}_{sdf}$ & \xmark & \xmark & 0.0371 & 0.0179 & 9.45 & \textbf{7.87} \\
\cmark & \cmark & \xmark & \xmark & 0.0361 & 0.0167 & 8.20 & 35.74 \\
\cmark & \cmark & \xmark & $\mathcal{L}_{t}$ & 0.0361 & 0.0167 & 8.40 & 35.74 \\
\cmark & \cmark & \cmark & $\mathcal{L}_{t}$ & 0.0368 & 0.0167 & 8.20 & 21.29 \\
\cmark & \cmark\!\!\cmark & \cmark & $\mathcal{L}_{t}$ & 0.2206 & 0.2082 & \textbf{3.63} & 31.60 \\
\cmark & \cmark & \cmark & \cmark & 0.0368 & 0.0174 & \underline{7.99} & \underline{10.66} \\
\midrule
\multicolumn{8}{c}{\textbf{Ablation study on vertex number for $\mathcal{L}_{lp}$}}\\
\midrule
\multicolumn{4}{l|}{4{,}000 (Ours)} & 0.0368 & 0.0174 & \underline{7.99} & \underline{10.66} \\
\multicolumn{4}{l|}{2{,}000} & \textbf{0.0355} & \textbf{0.0162} & 8.41 & 8.66 \\
\multicolumn{4}{l|}{1{,}000} & 0.0356 & 0.0163 & 8.38 & 8.55 \\
\multicolumn{4}{l|}{500} & 0.0378 & 0.0190 & 8.49 & 8.66 \\
\bottomrule[2pt] 
\end{tabular}
}
\label{tab:ablation}
\end{table}

We conduct extensive experiments to evaluate the significance of each proposed component, and the results are shown in the first part of \cref{tab:ablation}. The baseline follows R${}^2$ET \cite{r2et}, which applies an SDF loss for training but retargets motion frame-by-frame without considering temporal dependencies. Introducing the spatio-temporal (ST) model significantly reduces curvature, demonstrating the effectiveness of temporal modeling in reducing motion jitter. However, the penetration rate increases slightly.
Adding the limb penetration constraint ($\mathcal{L}_{lp}$) effectively mitigates interpenetration, reducing the penetration rate from $9.45\%$ to $8.20\%$. Further incorporating the basic smoothness loss ($\mathcal{L}_t$) improves temporal consistency without affecting penetration.  

Next, we analyze the impact of dense shape representation (DSR). Adding DSR alongside $\mathcal{L}_{lp}$ and $\mathcal{L}_t$ further reduces curvature (from 35.74 to 21.29), indicating that shape-aware features help the model predict smoother motion. However, an excessive penalty on penetration, such as applying a large $\mathcal{L}_{lp}$, reduces penetration significantly (to $3.63\%$) but leads to poor motion generalization, as reflected by the sharp increase in MSE and local MSE. Our full model, which balances spatial, temporal, and geometric constraints, achieves a pretty low penetration rate ($7.99\%$) while maintaining low curvature (10.66), demonstrating its overall effectiveness.

In the second part of \cref{tab:ablation}, we examine the effect of varying the number of sampled vertices for the limb penetration constraint. Reducing the number of sampled vertices from 4,000 (ours) to 500 slightly increases penetration and curvature, but the differences remain minimal, suggesting that our method is robust to different sampling densities. Considering the balance between computational efficiency and accuracy, we choose 4,000 vertices in our final model.
We conduct further ablation experiments for other modules, and please refer to the supplemental materials.

\section{Conclusion}

In this paper, we aim to achieve three key objectives in motion retargeting: semantics preservation, geometric plausibility, and temporal consistency. We propose STaR, a novel spatio-temporal skinned motion retargeting framework, embedded with a limb penetration constraint and a temporal consistency constraint. These constraints modify the motion manifold for characters of different shapes, while our spatio-temporal model design helps the optimizer find better parameters for this task. Experimental results demonstrate that our STaR achieves a superior balance across all three objectives compared to previous methods.

\noindent \textbf{Limitations and Future Work.} LBS is a major contributor to interpenetration. Resolving skinning artifacts could significantly lower penetration rates, since some issues stem from rigid-body assumptions that real skin deformation would naturally avoid.
We also do not tackle missing contact in this paper, due to frequent false positives.
Addressing these limitations will be a focus of our future research.

\section*{Acknowledgements}

GS acknowledges support by Engineering and Physical Sciences Research Council [grant number EP/Y009800/1], through Keystone project funding from Responsible AI UK (KP0016). This research utilized Queen Mary's Andrena HPC facility, supported by QMUL Research-IT.

{
    \small
    \bibliographystyle{ieeenat_fullname}
    \bibliography{main}
}

\newpage

\maketitlesupplementary
\renewcommand{\thesection}{\Alph{section}}
\setcounter{section}{0}

We include the following sections in the supplemental materials:
\begin{itemize}
\item Further detail of limb penetration loss.
\item Further detail of the evaluation metrics.
\item Real-world applications: retargeting motion from real humans.
\item User study.
\item Ablation study on joint orientation loss.
\item Ablation study on global motion prediction.
\item Single-pass motion retargeting and separate motion retargeting.
\item Ablation study on Dense Shape Representation.
\item Efficiency of limb penetration constraint module.
\item Demo videos.
\end{itemize}

\section{Further Detail of Limb Penetration Loss} \label{penetration}

\begin{figure}[ht]
  \centering
   \includegraphics[width=1.0\linewidth]{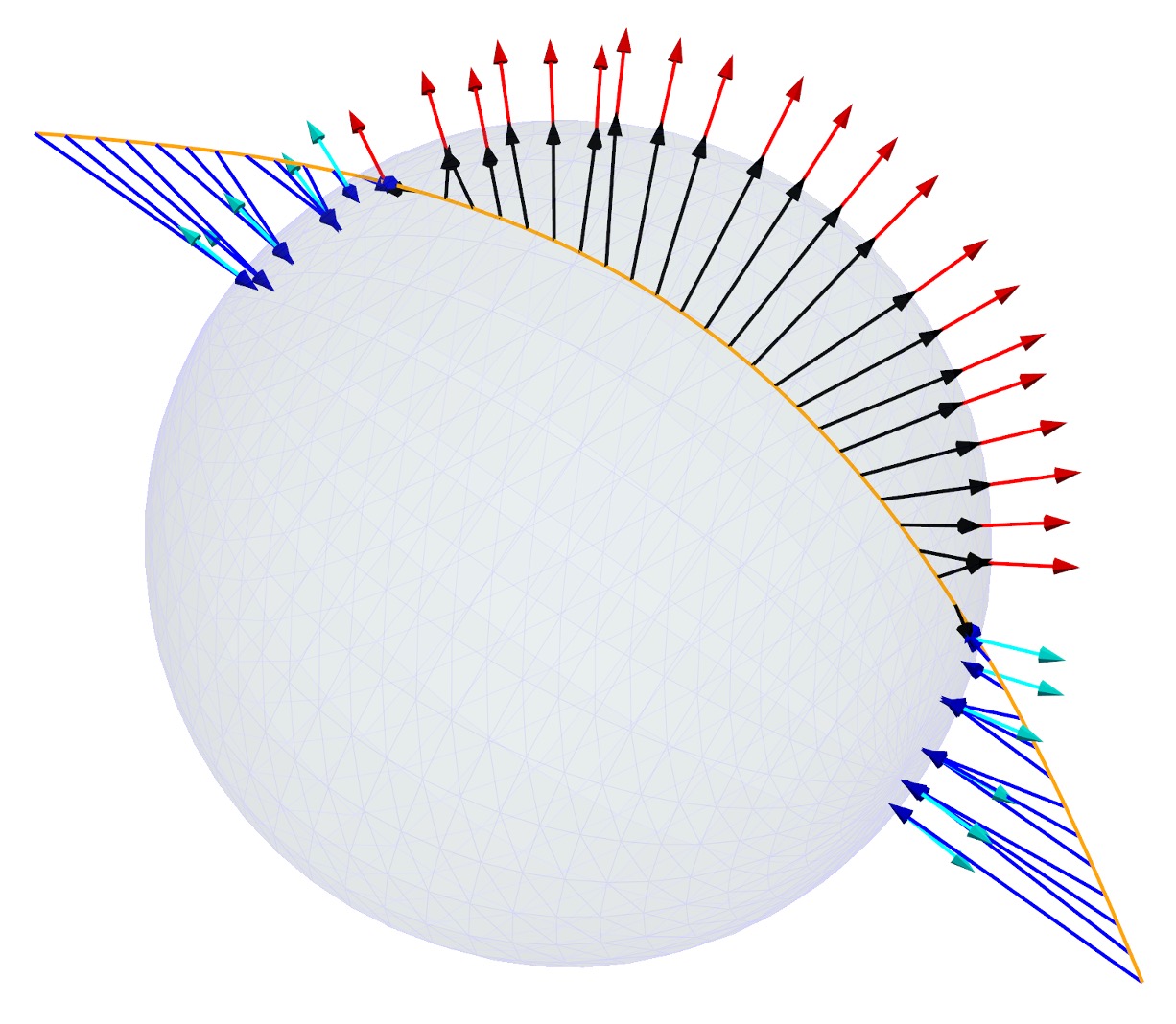}
   \caption{The detail of limb penetration loss computation for penetrated vertices and non-penetrating vertices.}
   \label{fig:mesh_ball}
\end{figure}

\Cref{fig:mesh_ball} illustrates penetration detection using Signed Distance Fields (SDF) in motion retargeting. The large sphere represents the body surface, while the curved trajectory represents the motion path of a limb or body part. Along this trajectory, multiple vertices exist, some penetrating the body (inside the sphere) and others remaining outside. The penetration loss computation involves finding the nearest reference vertex on the body surface for each motion path vertex, computing the vector from the query vertex to its reference vertex, and multiplying this vector by the normal vector of the reference vertex to estimate penetration depth. The black and red arrows represent surface normals, which guide penetration correction. Blue arrows indicate vectors from penetrating vertices to their nearest reference points, while cyan arrows represent similar vectors for non-penetrating vertices. This visualization highlights how full-body geometric correction is applied across the motion trajectory, ensuring that motion retargeting maintains geometric plausibility by preventing unnatural interpenetration.

\section{Evaluation Metrics} \label{metrics}

We evaluate the retargeted motion primarily from three perspectives: semantics, geometry, and motion smoothness. These are measured using MSE, penetration rate, and curvature, respectively.

\noindent \textbf{Mean Squared Error}. The Mean Squared Error (MSE) evaluates semantic preservation by assessing how closely the retargeted skeleton joints $\hat{\mathbf{X}}$, align with the ground truth joints, $\mathbf{X}_{gt}$. Although the ground truth suffers severe penetration, we investigate the motion sequences, and they can still work as an auxiliary evaluation of how semantics is maintained. The squared error is normalized by the character's height $h$. The metric is formulated as:

\begin{equation} \label{app_eq1}
\mathbf{MSE} = \frac{1}{h} \lVert \mathbf{X}_{gt} - \hat{\mathbf{X}} \rVert_2^2.
\end{equation}

\noindent \textbf{Penetration Rate}. The penetration rate is calculated as the ratio of interpenetrating points to the total number of limb vertices. Unlike \cite{r2et}, our approach considers all limbs:

\begin{equation} \label{app_eq2}
\mathbf{Pen\ Rate} = \frac{\text{Number of penetrated limb vertices}}{\text{Number of all limb vertices}} \times 100\%.
\end{equation}

\noindent \textbf{Curvature}. To address the discontinuity in the retargeted motion path, we compute the curvature of the motion path for each joint based on acceleration. Let $\mathbf{r}$ represent the motion vector, the curvature is defined as:

\begin{equation} \label{app_eq3}
\mathbf{Curv} = \norm{\dfrac{\mathrm{d}^2 \mathbf{r}}{\mathrm{d} t^2}}^2_2.
\end{equation}

\section{Real-world Applications} \label{realhuman}

\begin{figure}[ht]
  \centering
   \includegraphics[width=1.0\linewidth]{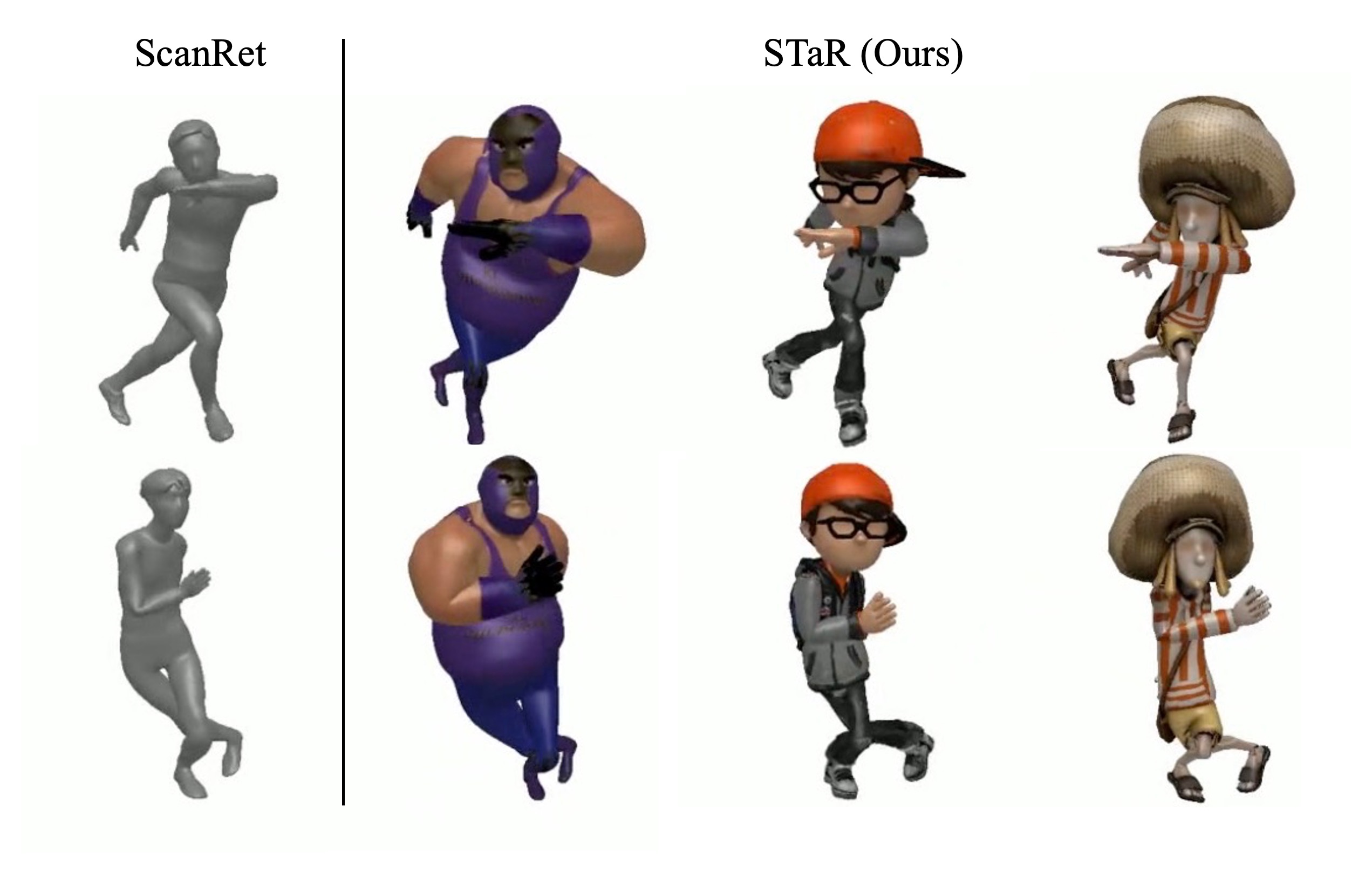}
   \caption{Motion retargeting from real human actors.}
   \label{fig:realworld}
\end{figure}

\Cref{fig:realworld} presents the motion retargeting results of our STaR on real-human motion data from the ScanRet dataset \cite{meshret}. The left column shows real-human motions with texture maps removed for anonymity, while the right columns display the retargeted motions on diverse target characters. STaR effectively preserves motion semantics, transferring poses and movement dynamics while adapting to different body shapes, including a wrestler, a stylized boy, and a cartoon-like figure. Despite variations in skeletal structures, STaR maintains spatial and temporal coherence, ensuring natural adaptation without excessive limb stretching or severe interpenetration. The results demonstrate STaR’s ability to generalize human motion to stylized characters while preserving geometric plausibility and temporal consistency.

\section{User Study} \label{userstudy}

\begin{table*}[htbp!]
\caption{Human evaluation results between our STaR and baseline methods.} 
\centering 
\begin{tabular}{l|cccc} 
\toprule[2pt] 
\textbf{Criteria} & SAN\cite{san} & R${}^2$ET \cite{r2et} & MeshRet\cite{meshret} & STaR (Ours) \\ 
\midrule 
\textbf{Semantics Preservation} & $11.43\%$ & $8.57\%$ & $2.86\%$ & $77.14\%$ \\
\textbf{Geometry Correctness}   & $12.38\%$ & $9.52\%$ & $4.76\%$ & $73.33\%$ \\
\textbf{Motion Smoothness}      & $20.00\%$ & $4.76\%$ & $4.76\%$ & $70.48\%$ \\
\textbf{Overall Quality}        & $12.38\%$ & $8.57\%$ & $2.86\%$ & $76.19\%$ \\
\bottomrule[2pt] 
\end{tabular}
\label{tab:userstudy}
\end{table*}

We conducted a user study to compare the retargeted motion sequences of STaR with three methods: SAN \cite{san}, R${}^{2}$ET \cite{r2et}, and MeshRet \cite{meshret}. Participants were presented with 10 randomly selected motion sequences, each containing the source motion and retargeted results from all four methods, shown in a randomized order to prevent bias.

Participants were asked to evaluate the results based on four key aspects:
\begin{enumerate}
\item Which one better preserves the semantics of the original motion?
\item Which one is physically more reasonable (fewer penetrations and distortions)?
\item Which movement appears smoother and more natural?
\item Which one do you prefer overall?
\end{enumerate}

To ensure a diverse set of responses, we distributed questionnaires and collected feedback from 21 participants. The results, presented in \cref{tab:userstudy}, show that STaR significantly outperforms all baseline methods across all criteria. Notably, $77.14\%$ of participants favored STaR in terms of motion semantics preservation, and $76.19\%$ preferred our results overall. Our model also received $70.48\%$ approval for motion coherence, demonstrating its ability to produce smooth and temporally stable motion trajectories, while achieving $73.33\%$ in geometric correctness, highlighting its effectiveness in preventing interpenetration and ensuring spatial plausibility.

Compared to other methods, SAN \cite{san} and R${}^{2}$ET \cite{r2et} struggle with geometric plausibility, while MeshRet \cite{meshret} fails on characters with diverse body shapes, leading to excessive penetration issues. In contrast, STaR consistently maintains a balance between motion semantics, geometric correctness, and temporal consistency, making it the preferred choice for high-quality motion retargeting.




\section{Ablation Study on Joint Orientation Loss} \label{joint}

\begin{figure}[ht]
  \centering
   \includegraphics[width=1.0\linewidth]{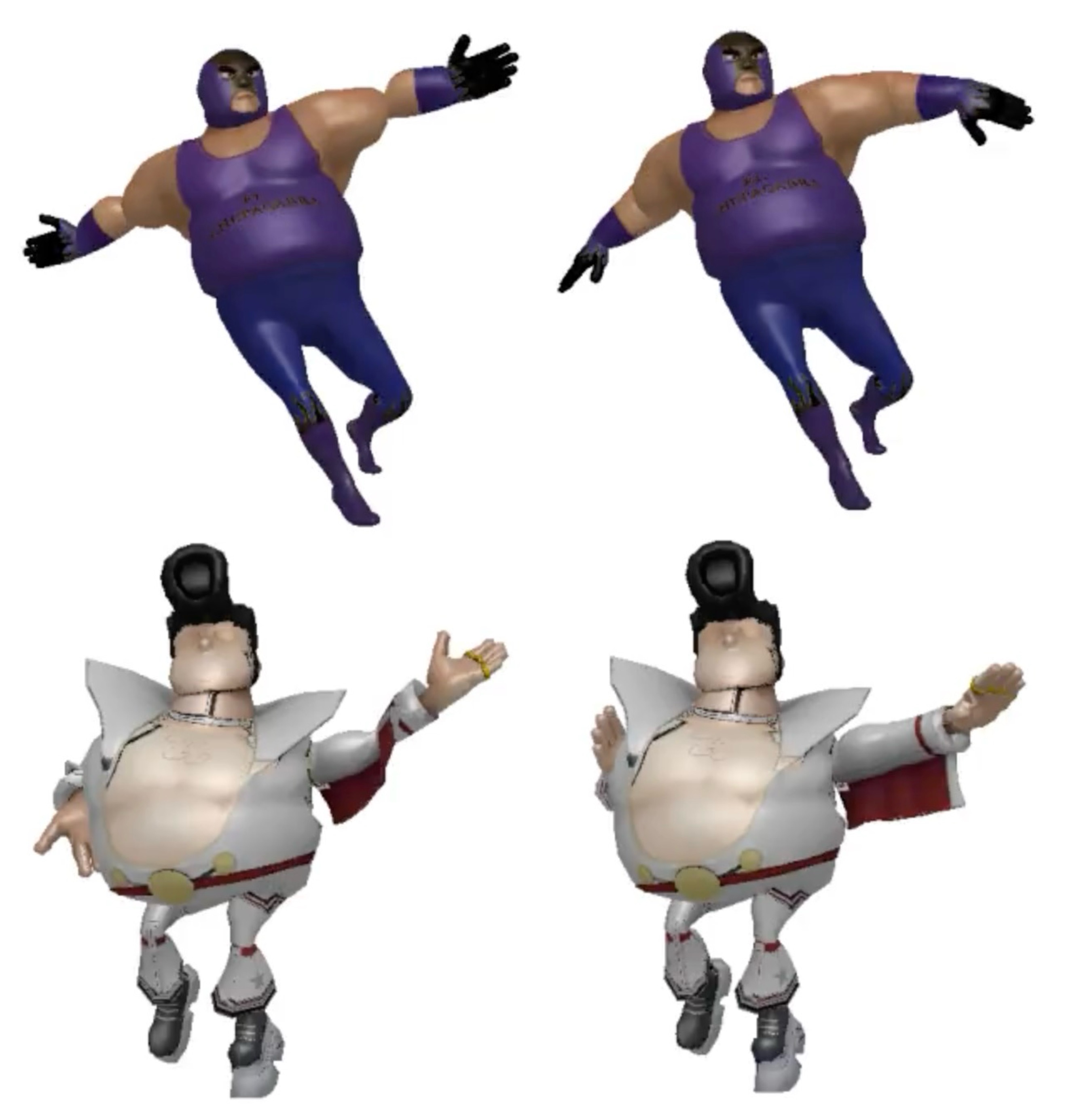}
   \caption{Ablation study on joint orientation loss. Without this loss (left), STaR's large search space may lead to unnatural poses, such as flipped arms. Adding the loss (right) ensures physically plausible joint orientations.}
   \label{fig:joint_orientation}
\end{figure}

\Cref{fig:joint_orientation} illustrates an ablation study on the joint orientation loss, comparing results without the loss (left) and with the loss (right). In the absence of this constraint, our STaR model, which operates in a large search space, occasionally produces unnatural poses, such as flipped arms or misaligned limb orientations. These artifacts arise due to the increased flexibility of the model, which, without explicit regularization, may lead to implausible joint rotations.

By incorporating the joint orientation loss, as shown on the right, the model effectively regulates joint rotations, ensuring physically plausible limb orientations while preserving motion semantics. This demonstrates the importance of enforcing orientation constraints to prevent extreme limb deviations and enhance the stability of motion retargeting.

\section{Ablation Study on Global Motion Prediction}

After examining the relationship between global motion and character height in the Mixamo dataset \cite{mixamo}, we observed no statistically significant correlation. After introducing a compact decoder to enhance global motion prediction, we observed rapid overfitting to the training set, resulting in poor generalization. As shown in \cref{tab:ablation_global}, its performance does not exceed the baseline, which simply normalizes and denormalizes global motion relative to character height.

\begin{table}[h]
\caption{Ablation study on global motion prediction. Since the global motion prediction module will only affect the \textbf{MSE} metric, the other three metrics are omitted for clarity. Please note that this study evaluates a preliminary variant—not our final model configuration.} 
\centering 
\scalebox{0.78}{
\begin{tabular}{l|cc|c|c} 
\toprule[2pt] 
\textbf{Methods} & $\ \textbf{MSE}\downarrow\ $ & $\textbf{MSE}^{lc}\downarrow$ & $\textbf{Pen}\%\downarrow$ & $\textbf{Curv}$ \rotatebox[origin=c]{180}{$\Lsh$} \\ 
\midrule 
\multicolumn{5}{c}{\textbf{Global Motion Prediction}}\\
\midrule
Baseline & \underline{1.7770} & - & - & - \\
Global Motion Decoder & 1.8708 & - & - & - \\
\bottomrule[2pt] 
\end{tabular}
}
\label{tab:ablation_global}
\end{table}

\section{Single-pass Motion Retargeting and Separate Motion Retargeting}

As presented in \cref{tab:ablation_sep}, our well-designed spatio-temporal model supports motion retargeting of varying sequence lengths within a single forward pass. The results show no significant difference between separate retargeting and single-pass retargeting.

\begin{table}[h]
\caption{The results of single-pass motion retargeting and separate motion retargeting} 
\centering 
\scalebox{0.77}{
\begin{tabular}{l|cc|c|c} 
\toprule[2pt] 
\textbf{Methods} & $\ \textbf{MSE}\downarrow\ $ & $\textbf{MSE}^{lc}\downarrow$ & $\textbf{Pen}\%\downarrow$ & $\textbf{Curv}$ \rotatebox[origin=c]{180}{$\Lsh$} \\ 
\midrule 
\multicolumn{5}{c}{\textbf{Separated Inference V.S. Inference Once}}\\
\midrule
Model 1 Separate Inference & 0.0369 & 0.0175 & 7.99 & 10.61 \\
Model 1 Inference Once & 0.0368 & 0.0174 & 7.99 & 10.66 \\
\midrule
Model 2 Separate Inference & 0.0355 & 0.0162 & 8.41 & 8.69 \\
Model 2 Inference Once & 0.0355 & 0.0162 & 8.41 & 8.66 \\
\bottomrule[2pt] 
\end{tabular}
}
\label{tab:ablation_sep}
\end{table}

\section{Ablation study on Dense Shape Representation}

For shape representation, we exclude skeleton data and skeleton-based shape information due to their limited utility and potential drawbacks. The skeleton bounding box's dimensions \cite{r2et} provide limited shape details, which are insufficient for effective motion retargeting that minimizes penetration. This limitation arises because motion is influenced by the shapes of limbs and other body parts. We show 3 results as proof in \cref{tab:ablation_dsr}: (1) skeleton data only; (2) our DSR; (3) their combination. Comparing (1) and (2), skeleton data is inadequate for preventing penetration. Additionally, (3) shows that integrating skeleton data with point clouds does not enhance results.

In DSR, the point cloud is NOT separated before being fed into the point cloud transformer \cite{pct}, and the $K$-channel shape representation is derived via a compact MLP from the comprehensive geometric feature produced by the transformer. Unlike frame-by-frame methods \cite{r2et, smt} that rely on spatially aligned skeleton data, our approach extracts both global and local information for each joint. This enables the spatial and temporal transformers to access comprehensive shape information, particularly benefiting the temporal network. In \cref{tab:ablation_dsr} (4), we show the result of extracting a separate point cloud for each joint as proof. Comparing (2) and (4), separating the points does not benefit the retargeting pipeline.

\begin{table}[h]
\caption{Ablation study on Dense Shape Representation.} 
\centering 
\scalebox{0.79}{
\begin{tabular}{l|cc|c|c} 
\toprule[2pt] 
\textbf{Methods} & $\ \textbf{MSE}\downarrow\ $ & $\textbf{MSE}^{lc}\downarrow$ & $\textbf{Pen}\%\downarrow$ & $\textbf{Curv}$ \rotatebox[origin=c]{180}{$\Lsh$} \\ 
\midrule 
\multicolumn{5}{c}{\textbf{Global Motion Prediction}}\\
\midrule
(1) Skeleton Info & 0.0367 & 0.0175 & 9.42 & 10.11 \\
(2) DSR (Ours) & 0.0368 & 0.0174 & \textbf{7.99} & 10.66 \\
(3) Skeleton Info + DSR & 0.0366 & 0.0174 & 8.22 & 10.05 \\
(4) Separate Point Cloud & 0.0934 & 0.0783 & 13.45 & \textbf{8.70} \\
\bottomrule[2pt] 
\end{tabular}
}
\label{tab:ablation_dsr}
\end{table}

\section{Efficiency of the Limb Penetration Constraint Module}

In \cref{tab:ablation_speed}, we compare the training speeds of (1) the SDF loss $\mathcal{L}_{sdf}$ from R${}^2$ET \cite{r2et}, (2) modified SDF loss \cite{r2et} including limb-limb penetration, and (3) our limb penetration constraint $\mathcal{L}_{lp}$. Using the same number of vertices, we measured the time per iteration on an RTX 4090 graphics card. Our limb penetration loss is approximately 8 times faster compared with the SDF loss from \cite{r2et}.

\begin{table}[h]
\caption{Loss Efficiency.} 
\centering 
\begin{tabular}{l|c}
\toprule[2pt]
\textbf{Methods} & $\ \textbf{Speed (s/iter)}\downarrow\ $ \\
\midrule
$\mathcal{L}_{sdf}$ \cite{r2et} & 127.66 \\
$\mathcal{L}_{sdf}$ \cite{r2et} w/ limb & 133.77 \\
$\mathcal{L}_{lp}$ (Ours) & \textbf{16.32} \\
\bottomrule[2pt]
\end{tabular}
\label{tab:ablation_speed}
\end{table}

\section{Demo Videos}

We provide demo videos to showcase the performance of our STaR method. These videos show the motion retargeting from real-human datasets, ScanRet \cite{meshret}. We include the demo videos in the supplementary materials.

\end{document}